\pdfoutput=1

\documentclass[11pt]{article}

\usepackage{emnlp2021}

\usepackage{times}
\usepackage{latexsym}

\usepackage[T1]{fontenc}

\usepackage[utf8]{inputenc}

\usepackage{microtype}

\usepackage{setspace}

\usepackage{smile}
\usepackage{hhline}
\usepackage{multirow}
\usepackage{mathtools}
\usepackage[ruled,vlined]{algorithm2e}
\usepackage{subcaption}
\usepackage{hyperref}

\newcommand{\dd}{\mathrm{d}}

\title{Adversarial Regularization as Stackelberg Game: \\An Unrolled Optimization Approach}

\author{
Simiao Zuo$^\ddagger$\thanks{\hspace{0.03in} Corresponding author.}, \ Chen Liang$^\ddagger$, \ Haoming Jiang$^\Box$\thanks{\hspace{0.03in} Work was done at Georgia Institute of Technology.}, \
Xiaodong Liu$^\diamond$, \ Pengcheng He$^\diamond$, \\
\textbf{Jianfeng Gao$^\diamond$, \ Weizhu Chen$^\diamond$ and Tuo Zhao$^{\ddagger}$} \\
$^\ddagger$Georgia Institute of Technology \ \
$^\Box$Amazon \ \
$^\diamond$Microsoft \\
\texttt{\{simiaozuo,cliang73\}@gatech.edu}, \ \texttt{jhaoming@amazon.com} \\
\texttt{\{xiaodl,Pengcheng.H,jfgao,wzchen\}@microsoft.com}, \\ \texttt{tourzhao@gatech.edu}
}

\begin{document}
\maketitle

\begin{abstract}
Adversarial regularization has been shown to improve the generalization performance of deep learning models in various natural language processing tasks. Existing works usually formulate the method as a zero-sum game, which is solved by alternating gradient descent/ascent algorithms. Such a formulation treats the adversarial and the defending players equally, which is undesirable because only the defending player contributes to the generalization performance. To address this issue, we propose Stackelberg Adversarial Regularization (SALT), which formulates adversarial regularization as a Stackelberg game. This formulation induces a competition between a leader and a follower, where the follower generates perturbations, and the leader trains the model subject to the perturbations. Different from conventional approaches, in SALT, the leader is in an advantageous position. When the leader moves, it recognizes the strategy of the follower and takes the anticipated follower's outcomes into consideration. Such a leader's advantage enables us to improve the model fitting to the unperturbed data. The leader's strategic information is captured by the Stackelberg gradient, which is obtained using an unrolling algorithm. Our experimental results on a set of machine translation and natural language understanding tasks show that SALT outperforms existing adversarial regularization baselines across all tasks.
Our code is available at \url{https://github.com/SimiaoZuo/Stackelberg-Adv}.
\end{abstract}

\section{Introduction}

Adversarial regularization \cite{miyato2016adversarial} has been shown to improve the generalization performance of deep learning models in various natural language processing (NLP) tasks, such as language modeling \cite{wang2019improving}, machine translation \cite{sano2019effective}, natural language understanding \cite{jiang2019smart}, and reading comprehension \cite{jia2017adversarial}.
However, even though significant progress has been made, the power of adversarial regularization is not fully harnessed.

Conventional adversarial regularization is formulated as a zero-sum game (a min-max optimization problem), where two players seek to minimize/maximize their utility functions.
In this formulation, an adversarial player composes perturbations, and a defending player solves for the model parameters subject to the perturbed inputs.
Existing algorithms find the equilibrium of this zero-sum game using alternating gradient descent/ascent \cite{madry2017towards}.
For example, in a classification problem, the adversarial player first generates the input perturbations by running projected gradient ascent to maximize a loss function, and then the defending player updates the model using gradient descent, trying to decrease the classification error.
Notice that in this case, neither of the players know the strategy of its competitor, i.e., the model does not know how the perturbations are generated, and vice versa.
In other words, the two players are of the same priority, and either one of them can be advantageous in the game. It is possible that the adversarial player generates over-strong perturbations that hinder generalization of the model.


To resolve this issue, we grant the defending player (i.e., the model) a higher priority than the adversarial player by letting the defender recognize its competitor's strategy, such that it is advantageous in the game.
Consequently, we propose \textbf{S}tackelberg \textbf{A}dversaria\textbf{l} Regulariza\textbf{t}ion (SALT), where we formulate adversarial regularization as a Stackelberg game \cite{von2010market}. The concept arises from economics, where two firms are competing in a market, and one of the them is in the leading position by acknowledging the opponent's strategy.
In Stackelberg adversarial regularization, a leader solves for the model parameters, and a follower generates input perturbations.
The leader procures its advantage by considering what the best response of the follower is, i.e., how will the follower respond after observing the leader's decision. Then, the leader minimizes its loss, anticipating the predicted response of the follower.

The SALT framework identifies the interaction between the leader and the follower by treating the follower's strategy (i.e., the input perturbations) as an operator of the leader's decision (i.e., the model parameters). Then we can solve for the model parameters using gradient descent.
One caveat is that computing the gradient term, which we call the Stackelberg gradient, requires differentiating the interaction operator.
To rigorously define this operator, recall that the follower can be approximately solved using gradient ascent. We can treat the perturbations in each iteration as an operator of the model parameters, and the interaction operator is then the composition of such update-induced operators.
Correspondingly, the Stackelberg gradient is obtained by differentiating through these updates.
This procedure is referred to as unrolling \cite{pearlmutter2008reverse}, and the only computational overhead caused by it is computing Hessian vector products. As a result, when applying the finite difference method, computing the Stackelberg gradient requires two backpropagation and an extra $O(d)$ complexity operation, where $d$ is the embedding dimension.
Therefore, the unrolling algorithm computes the Stackelberg gradient without causing much computational overhead.

We conduct experiments on neural machine translation (NMT) and natural language understanding (NLU) tasks.
For the NMT tasks, we experiment on four low-resource and one rich-resource datasets. SALT improves upon existing adversarial regularization algorithms by notable margins, especially on low-resource datasets, where it achieves up to 2 BLEU score improvements.
To test performance on NLU tasks, we evaluate SALT on the GLUE \cite{wang2018glue} benchmark. SALT outperforms state-of-the-art models, such as BERT \cite{devlin2018bert},  FreeAT \cite{shafahi2019adversarial}, FreeLB \cite{zhu2019freelb}, and SMART \cite{jiang2019smart}. We build SALT on the BERT-base architecture, and we achieve an average score of 84.5 on the GLUE development set, which is at least 0.7 higher than existing methods. Moreover, even though we adapt SALT to BERT-base, the performance is noticeably higher than the vanilla BERT-large model (84.5 vs. 84.0).

The unrolling procedure was first proposed for auto-differentiation \cite{pearlmutter2008reverse}, and later applied in various context, such as hyper-parameter optimization \cite{maclaurin2015gradient, finn2017model}, meta-learning \cite{andrychowicz2016learning}, and Generative Adversarial Networks \cite{metz2016unrolled}.
To the best of our knowledge, we are the first to apply the unrolling technique to adversarial regularization to improve generalization performance.

We summarize our contributions as the following:
(1) We propose SALT, which employs a Stackelberg game formulation of adversarial regularization.
(2) We use an unrolling algorithm to find the equilibrium of the Stackelberg game.
(3) Extensive experiments on NMT and NLU tasks verify the efficacy of our method.

\vspace{0.05in}
\noindent
\textbf{Notation.}
We use $\dd f(x) / \dd x$ to denote the gradient of $f$ with respect to $x$. We use $\partial f(x,y) / \partial x$ to denote the partial derivative of $f$ with respect to $x$.
For a $d$-dimensional vector $v$, its $\ell_2$ norm is defined as $\norm{v}_2 = (\sum_{i=1}^d v_i^2)^{1/2}$, and its $\ell_\infty$ norm is defined as $\norm{v}_\infty = \max_{1 \leq i \leq d} |v_i|$.

\section{Background and Related Works}

\noindent $\diamond$
\textbf{Neural machine translation} has achieved superior empirical performance \cite{bahdanau2014neural, gehring2017convolutional, vaswani2017attention}. We focus on the Transformer architecture \cite{vaswani2017attention}, which integrates the attention mechanism in an encoder-decoder structure.
The encoder in a Transformer model first maps a source sentence into an embedding space, then the embeddings are fed into several encoding layers to generate hidden representations, where each of the encoding layers contains a self-attention mechanism and a feed-forward neural network (FFN). After which the Transformer decoder layers, each contains a self-attention, a encoder-decoder attention, and a FFN, decode the hidden representations.

\vspace{0.1in}
\noindent $\diamond$
\textbf{Adversarial training} was originally proposed for training adversarial robust classifiers in image classification \cite{szegedy2013intriguing, goodfellow2014explaining, madry2017towards}.
The idea is to synthesize strong adversarial samples, and the classifier is trained to be robust to them.
Theoretical understanding \cite{li2019inductive} about adversarial training and various algorithms to generate the adversarial samples, such as learning-to-learn \cite{jiang2021learning}, are proposed.
Besides computer vision, adversarial training can also benefit reinforcement learning \cite{shen2020deep}.
Different from the above fields, in NLP, the goal of adversarial training is to build models that generalize well on the unperturbed test data.
Note that robustness and generalization are different concepts. Recent works \cite{raghunathan2020understanding, min2020curious} showed that adversarial training can hurt generalization performance, i.e., accuracy on clean data.
As such, adversarial training needs to be treated with great caution. Therefore, in NLP, this technique requires refined tuning of, for example, the training algorithm and the perturbation strength.

\vspace{0.1in}
\noindent $\diamond$
\textbf{Fine-tuning pre-trained language models} \cite{peters2018deep, devlin2018bert, radford2019language, liu2019roberta, he2020deberta} is state-of-the-art for natural language understanding tasks such as the GLUE \cite{wang2018glue} benchmark.
Recently, there are works that use adversarial pre-training \cite{liu2020adversarial} and adversarial-regularized fine-tuning methods such as SMART \cite{jiang2019smart}, FreeLB \cite{zhu2019freelb}, and FreeAT \cite{shafahi2019adversarial} to improve model generalization and robustness \cite{cheng2020posterior}.

\section{Method}

Natural language inputs are discrete symbols (e.g., words), instead of continuous ones. Therefore, a common approach to generate perturbations is to learn continuous embeddings of the inputs and operate on the embedding space \cite{miyato2016adversarial, clark2018semi, sato2018interpretable, sano2019effective, stutz2019disentangling}.
Let $f(x,\theta)$ be our model, where $x$ is the input embedding, and $\theta$ is the model parameter. Further let $y$ be the ground-truth output corresponding to $x$.
For example, in NMT, $f$ is a sequence-to-sequence model, $x$ is the embedding of the source sentence, and $y$ is the target sentence. In classification tasks, $f$ is a classifier, $x$ is the input sentence/document embedding, and $y$ is the label.
In both of these cases, the model is trained by minimizing the empirical risk over the training data, i.e.,
\begin{equation*}
    \min_{\theta} \cL(\theta)=\frac{1}{n} \sum_{i=1}^n \ell( f(x_i,\theta), y_i ).
\end{equation*}
Here $\{(x_i,y_i)\}_{i=1}^n$ is our dataset, and $\ell$ is a task-specific loss function, e.g., cross-entropy loss.

\subsection{Adversarial Regularization}

Adversarial Regularization \cite{miyato2016adversarial} is a regularization technique that encourages smoothness of the model outputs around each input data point.
Concretely, we define an adversarial regularizer for non-regression tasks as
\begin{align*}
    &\ell_{v}(x, \delta, \theta) = \mathrm{KL}\big( f(x, \theta) ~||~ f(x + \delta, \theta) \big), \\[5pt]
    &\text{where } \mathrm{KL}(P ~||~ Q) = \sum_k p_k \log \frac{p_k}{q_k}.
\end{align*}
Here $\mathrm{KL}(\cdot || \cdot)$ is the Kullback–Leibler (KL) divergence, $\delta$ is the perturbation corresponding to $x$, and $f(\cdot, \theta)$ is the prediction probability simplex given model parameters $\theta$.
In regression tasks, the model output $f(\cdot,\theta)$ is a scalar, and the adversarial regularizer is defined as
\begin{equation*}
    \ell_v \left(x, \delta, \theta) = (f(x,\theta) - f(x+\delta,\theta) \right)^2.
\end{equation*}
Then the training objective is
\begin{equation} \label{eq:vat-loss}
    \min_{\theta} \cL(\theta)
    + \frac{\alpha}{n} \sum_{i=1}^n\max_{\norm{\delta_i} \leq \epsilon} \ell_{v}( x_i, \delta_i, \theta ) ,
\end{equation}
where $\alpha$ is a tuning parameter, $\epsilon$ is a pre-defined perturbation strength, and $\norm{\cdot}$ is either the $\ell_2$ norm or the $\ell_\infty$ norm.

The min and max problems are solved using alternating gradient descent/ascent. We first generate the perturbations $\delta$ by solving the maximization problem using several steps of projected gradient ascent, and then we update the model parameters $\theta$ with gradient descent, subject to the perturbed inputs.
More details are deferred to Appendix~\ref{app:vat}.

One major drawback of the zero-sum game formulation (Eq.~\ref{eq:vat-loss}) is that it fails to consider the interaction between the perturbations $\delta$ and the model parameters $\theta$.
This is problematic because a small change in $\delta$ may lead to a significant change in $\theta$, which renders the optimization ill-conditioned. Thus, the model is susceptible to underfitting and generalize poorly on unperturbed test data.

\subsection{Adversarial Regularization as Stackelberg Game}

We formulate adversarial regularization as a Stackelberg game \cite{von2010market}:
\begin{align} \label{eq:bilevel}
    &\min_\theta \cF(\theta)=
    \cL(\theta) + \frac{\alpha}{n}\sum_{i=1}^n \ell_{v}(x_i, \delta_i^K(\theta), \theta ), \notag\\
    &\text{s.t. } \delta_i^K(\theta) = U^K \circ U^{K-1} \circ \cdots \circ U^1(\delta_i^0).
\end{align}
Here ``$\circ$'' denotes operator composition, i.e., $f \circ g (\cdot) = f(g(\cdot))$.
Following conventions, in this Stackelberg game, we call the optimization problem in Eq.~\ref{eq:bilevel} the leader. Further, the follower in Eq.~\ref{eq:bilevel} is described using a equality constraint.
Note that $U^K$ is the follower's $K$-step composite strategy, which is the composition of $K$ one-step strategies $\{U^k\}_{k=1}^K$. 
In practice, $K$ is usually small. This is because in NLP, we target for generalization, instead of robustness, and choosing a small $K$ prevents over-strong adversaries.

In Eq.~\ref{eq:bilevel}, $U^k$s are the follower's one-step strategies, and we call them update operators, e.g., $U^1$ updates $\delta^0$ to $\delta^1$ using pre-selected algorithms.
For example, projected gradient ascent can be applied as the update procedure, that is,
\begin{align} \label{eq:unroll-forward}
    \delta^k(\theta) &= U^k(\delta^{k-1}(\theta)) \notag\\[5pt]
    &= \Pi_{\norm{\cdot} \leq \epsilon} \bigg( \delta^{k-1}(\theta)
    + \eta \frac{\partial \ell_v( x, \delta^{k-1}(\theta), \theta) }{\partial \delta^{k-1}(\theta)} \bigg) \notag\\
    &\text{for } k = 1, \cdots, K,
\end{align}
where $\delta^0 \sim \cN(0,\sigma^2 \mathrm{I})$ is a initial random perturbation drawn from a normal distribution with variance $\sigma^2 I$, $\eta$ is a pre-defined step size, and $\Pi$ denotes projection to the $\ell_2$-ball or the $\ell_{\infty}$-ball.

To model how the follower will react to a leader's decision $\theta$, we consider the function $\delta^K(\theta)$. Then, adversarial training can be viewed solely in terms of the leader decision $\theta$.

We highlight that in our formulation, the leader knows the strategy, instead of only the outcome, of the follower. This information is captured by the Stackelberg gradient $\dd \cF(\theta) / \dd \theta$, defined as
the following:
\begin{align} \label{eq:stackelberg-gradient}
    \frac{\dd \cF(\theta)}{\dd \theta}
    &= \frac{\dd \ell( f(x, \theta), y )}{\dd \theta}
    + \alpha \frac{\dd \ell_{v}( x, \delta^K(\theta), \theta )}{\dd \theta} \nonumber\\
    &= \underbrace{
    \frac{\dd \ell( f(x, \theta), y )}{\dd \theta}
    + \alpha \frac{\partial \ell_{v}( x, \delta^K, \theta )}{\partial \theta}
    }_{\text{leader}} \nonumber\\[5pt]
    &+ \underbrace{\alpha \frac{\partial \ell_{v}( x, \delta^K(\theta), \theta )}{\partial \delta^K(\theta)}
    \frac{\dd \delta^K(\theta)}{\dd \theta}}_{\text{leader-follower interaction}}.
\end{align}
The underlying idea behind Eq.~\ref{eq:stackelberg-gradient}\footnote{The second term in ``leader'' is written as $\partial \ell_v(x, \delta^K, \theta) / \partial \theta$, instead of $\partial \ell_v(x, \delta^K(\theta), \theta) / \partial \theta$. This is because the partial derivative of $\theta$ is only taken w.r.t. the third argument in $\ell_v(x,\delta^K, \theta)$. We drop the $\theta$ in $\delta^K(\theta)$ to avoid causing any confusion.} is that given a leader's decision $\theta$, we take the follower’s strategy into account (i.e., the ``leader-follower interaction'' term) and find a direction along which the leader’s loss decreases the most. Then we update $\theta$ in that direction.
Note that the gradient used in standard adversarial training (Eq.~\ref{eq:vat-loss}) only contains the ``leader'' term, such that the ``leader-follower interaction'' is not taken into account.


\subsection{SALT: Stackelberg Adversarial Regularization}

\begin{algorithm}[!t]
\SetAlgoLined
\caption{Stackelberg Adversarial Regularization with Unrolled Optimization.}
\label{alg:unroll}
\KwIn{$\cD$: dataset; $T$: total number of training epochs; $\sigma^2$: variance of initial perturbations; $K$: number of unrolling steps; Optimizer: optimizer to update $\theta$.}
\textbf{Initialize:} model parameters $\theta$\;
\For{$t = 1, \cdots, T$}{
\For{$(x,y) \in \cD$}{
    Initialize $\delta^0 \sim \cN(0,\sigma^2 \mathrm{I})$\;
    \For{$k = 1, \cdots, K$}{
        Compute $\delta^k$ using Eq.~\ref{eq:unroll-forward}\;
        Compute $\dd \delta^k(\theta) / \dd \theta$ using Eq.~\ref{eq:unroll-backward}\;
    }
    Compute $\dd \cF(\theta) / \dd \theta$ based on $\dd \delta^K(\theta) / \dd \theta$ using Eq.~\ref{eq:stackelberg-gradient}\;
    $\theta \leftarrow \text{Optimizer}(\dd \cF(\theta) / \dd \theta)$\;
}
}
\KwOut{$\theta$}
\end{algorithm}

We propose to use an unrolling method \cite{pearlmutter2008reverse} to compute the Stackelberg gradient (Eq.~\ref{eq:stackelberg-gradient}).
The general idea is that since the interaction operator is defined as the composition of the $\{U^k\}$ operators, all of which are known, we can directly compute the derivative of $\delta^K(\theta)$ with respect to $\theta$. Concretely, we first run a forward iteration to update $\delta$, and then we differentiate through this update to acquire the Stackelberg gradient.



Note that the updates of $\delta$ can take any form, such as projected gradient ascent in Eq.~\ref{eq:unroll-forward}, or more complicated alternatives like Adam \cite{kingma2014adam}.
For notation simplicity, we denote $\Delta( x, \delta^{k-1}(\theta), \theta) =\delta^k(\theta) - \delta^{k-1}(\theta)$.
Accordingly, Eq.~\ref{eq:unroll-forward} can be rewritten as
\begin{align} \label{eq:unroll-forward-diff}
    &\delta^k(\theta) = \delta^{k-1}(\theta) + \Delta( x, \delta^{k-1}(\theta), \theta).
\end{align}


The most expensive part in computing the Stackelberg gradient (Eq.~\ref{eq:stackelberg-gradient}) is to calculate $\dd \delta^K(\theta) / \dd \theta$, which involves differentiating through the composition form of the follower's strategy:
\begin{align} \label{eq:unroll-backward}
    \frac{\dd \delta^k(\theta)}{\dd \theta}
    &= \frac{\dd \delta^{k-1}(\theta)}{\dd \theta} 
    + \frac{\partial \Delta(x, \delta^{k-1}, \theta)}{\partial \theta} \notag\\
    &+ \frac{\partial \Delta(x, \delta^{k-1}(\theta), \theta)}{\partial \delta^{k-1}(\theta)}
    \frac{\dd \delta^{k-1}(\theta)}{\dd \theta} \notag\\
    &\quad \text{for }k = 1, \cdots, K.
\end{align}


We can compute Eq.~\ref{eq:unroll-backward} efficiently using deep learning libraries, such as PyTorch \cite{paszke2019pytorch}. Notice that $\Delta(x, \delta^{k-1}(\theta), \theta)$ already contains the first order derivative with respect to the perturbations. Therefore, the term $\partial \Delta(x, \delta^{k-1}(\theta), \theta) / \partial \delta^{k-1}(\theta)$ contains the Hessian of $\delta^{k-1}(\theta)$.
As a result, in Eq.~\ref{eq:stackelberg-gradient}, the most expensive operation is the Hessian vector product (Hvp).
Using the finite difference method, computing Hvp only requires two backpropagation and an extra $O(d)$ complexity operation.
This indicates that in comparison with conventional adversarial training, SALT does not introduce significant computational overhead.
The training algorithm is summarized in Algorithm~\ref{alg:unroll}.

\section{Experiments}

In all the experiments, we use \textit{PyTorch}\footnote{\url{https://pytorch.org/}} \cite{paszke2019pytorch} as the backend. All the experiments are conducted on NVIDIA V100 32GB GPUs.
We use the \textit{Higher} package\footnote{\url{https://github.com/facebookresearch/higher}} \cite{grefenstette2019generalized} to implement the proposed algorithm.

\subsection{Baselines}

We adopt several baselines in the experiments.

\vspace{0.05in}
\noindent $\diamond$ \textit{Transformer} \cite{vaswani2017attention} achieves superior performance in neural machine translation.

\vspace{0.05in}
\noindent $\diamond$ \textit{BERT} \cite{devlin2018bert} is a pre-trained language model that exhibits outstanding performance after fine-tuned on downstream NLU tasks.

\vspace{0.05in}
\noindent $\diamond$ \textit{Adversarial training} (\textit{Adv}, \citealt{sano2019effective}) in NMT can improve models' generalization by training the model to defend against adversarial attacks.

\vspace{0.05in}
\noindent $\diamond$ \textit{FreeAT} \cite{shafahi2019adversarial} enables ``free'' adversarial training by recycling the gradient information generated when updating the model parameters. This method was proposed for computer vision tasks, but was later modified for NLU. We further adjust the algorithm for NMT tasks. 

\vspace{0.05in}
\noindent $\diamond$ \textit{FreeLB} \cite{zhu2019freelb} is a ``free'' large batch adversarial training method. We modify FreeLB to an adversarial regularization method that better fits our need.
This algorithm was originally proposed for NLU. We modify the algorithm so that it is also suitable for NMT tasks.

\vspace{0.05in}
\noindent $\diamond$ \textit{SMART} \cite{jiang2019smart} is a state-of-the-art fine-tuning method that utilizes smoothness-inducing regularization and Bregman proximal point optimization.

We highlight that we focus on \textbf{model generalization} on clean data, instead of adversarial robustness (a model's ability to defend adversarial attacks). As we will see in the experiments, adversarial training methods (e.g., Adv, FreeAT) suffer from label leakage, and do not generalize as well as adversarial regularization methods.

\subsection{Neural Machine Translation}

\noindent
\textbf{Datasets.}
We adopt three low-resource datasets and a rich-resource dataset. Dataset statistics are summarized in Table~\ref{tb:dataset}.
For the low-resource experiments, we use\footnote{\url{https://iwslt.org/}}: English-Vietnamese from IWSLT'15, German-English from IWSLT'14, and French-English from IWSLT'16.
For the rich-resource experiments, we use the English-German dataset from WMT'16, which contains about 4.5 million training samples.

\begin{table}[t!]
\centering
\begin{tabular}{c|cccc}
\toprule 
\textbf{Data}  & \textbf{Source} & \textbf{Train} & \textbf{Valid} & \textbf{Test} \\ \midrule
\textbf{En-Vi} & IWSLT'15        & 133k           & 768            & 1268          \\
\textbf{De-En} & IWSLT'14        & 161k           & 7.2k           & 6.7k          \\
\textbf{Fr-En} & IWSLT'16        & 224k           & 1080           & 1133          \\ \hline
\textbf{En-De} & WMT'16          & 4.5m           & 3.0k           & 3.0k          \\ 
\bottomrule
\end{tabular}
\vskip -0.05in
\caption{Dataset source and statistics. Here ``k'' stands for thousand, and ``m'' stands for million.}
\label{tb:dataset}
\end{table}

\newcolumntype{C}{@{\hskip4pt}c@{\hskip4pt}}
\begin{table}[t!]
\centering
\begin{tabular}{l|ccc}
    \toprule
    & \textbf{En-Vi} & \textbf{De-En} & \textbf{Fr-En}\\ \midrule
    Transformer & 30.3 & 34.7 & 38.2 \\ 
    Adv & 31.0 & 34.8 & 38.8 \\
    FreeAT & 31.0 & 35.2 & 38.6 \\
    FreeLB & 31.6 & 35.3 & 38.7 \\ 
    SMART & 31.5 & 35.5 & 38.9 \\ \midrule 
    SALT & \textbf{32.8} & \textbf{36.8} & \textbf{39.7} \\
    \bottomrule
\end{tabular}
\vspace{-0.05in}
\caption{BLEU score on three low-resource datasets. All the baseline results are from our re-implementation. We report the mean of three runs.}
\label{tb:low-resource}
\end{table}

\begin{table}[t!]
\centering
\begin{tabular}{l|c}
    \toprule
    & \textbf{BLEU} \\ \midrule
    Transformer \cite{vaswani2017attention} & 28.4 \\
    FreeAT \cite{shafahi2019adversarial} & 29.0 \\
    FreeLB \cite{zhu2019freelb} & 29.0 \\ 
    SMART \cite{jiang2019smart} & 29.1 \\ \midrule
    SALT & \textbf{29.6} \\ 
    \bottomrule
\end{tabular}
\vspace{-0.05in}
\caption{sacreBLEU score on WMT'16 En-De. All the baseline results are from our re-implementation.}
\label{tb:rich-resource}
\vspace{-0.05in}
\end{table}

\begin{table*}[t!]
\centering
\resizebox{1.0\textwidth}{!}{
\begin{tabular}{@{\hskip3pt}l@{\hskip3pt}|C|C|C|C|C|C|C|C|C}
\bottomrule
& \textbf{RTE} & \textbf{MRPC} & \textbf{CoLA} & \textbf{SST-2} & \textbf{STS-B} & \textbf{QNLI} & \textbf{QQP} & \textbf{MNLI-m/mm} & \textbf{Average} \\
& Acc & Acc/F1 & Mcc & Acc & P/S Corr & Acc & Acc/F1 & Acc & \textbf{Score} \\ \midrule
BERT\textsubscript{LARGE} & 71.1 & 86.0/89.6 & 61.8 & 93.5 & 89.6/89.3 & 92.4& 91.3/88.4 & 86.3/86.2 & 84.0 \\ \midrule 
BERT\textsubscript{BASE} & 63.5 & 84.1/89.0 & 54.7 & 92.9 & 89.2/88.8 & 91.1 & 90.9/88.3 & 84.5/84.4 & 81.5 \\
FreeAT & 68.0 & 85.0/89.2 & 57.5 & 93.2 & 89.5/89.0 & 91.3 & 91.2/88.5 & 84.9/85.0 & 82.6 \\
FreeLB & 70.0 & 86.0/90.0 & 58.9 & 93.4 & 89.7/89.2 & 91.5 &91.4/88.4& 85.4/85.5 & 83.3 \\
SMART & 71.2 & 87.7/91.3 & 59.1 & 93.0 & 90.0/89.4 & 91.7& 91.5/88.5 & 85.6/86.0 & 83.8 \\ \hline
SALT & \textbf{72.9} & \textbf{88.4/91.8} & \textbf{61.0} & \textbf{93.6} & \textbf{90.4/90.0} & \textbf{92.0} &\textbf{91.7/88.6} & \textbf{86.1/85.8} & \textbf{84.5} \\
\bottomrule
\end{tabular}
}
\vspace{-0.05in}
\caption{Evaluation results on the GLUE development set. All the rows use \textit{BERT\textsubscript{BASE}}, except the top one, which is included to demonstrate the effectiveness of our model.
Best results on each dataset, excluding \textit{BERT\textsubscript{LARGE}}, are shown in \textbf{bold}. Results of \textit{BERT\textsubscript{BASE}} \cite{devlin2018bert}, \textit{BERT\textsubscript{LARGE}} \cite{devlin2018bert}, \textit{FreeAT} \cite{shafahi2019adversarial}, and \textit{FreeLB} \cite{zhu2019freelb} are from our re-implementation. \textit{SMART} results are from \citet{jiang2019smart}.}
\label{tb:glue-results}
\end{table*}

\begin{table*}[t!]
\centering
\resizebox{1.0\textwidth}{!}{
\begin{tabular}{@{\hskip3pt}l@{\hskip3pt}|C|C|C|C|C|C|C|C|C}
\bottomrule
& \textbf{RTE} & \textbf{MRPC} & \textbf{CoLA} & \textbf{SST-2} & \textbf{STS-B} & \textbf{QNLI} & \textbf{QQP} & \textbf{MNLI-m/mm} & \textbf{Average} \\
& Acc & Acc/F1 & Mcc & Acc & P/S Corr & Acc & Acc/F1 & Acc & \textbf{Score} \\ \midrule
BERT\textsubscript{BASE} & 66.4 & 84.8/88.9 & 52.1 & 93.5 & 87.1/85.8 & 90.5 & 71.2/89.2 & 84.6/83.4 & 80.0 \\
FreeLB & 70.1 & 83.5/88.1 & 54.5 & 93.6 & 87.7/86.7 & 91.8 & 72.7/89.6 & 85.7/84.6 & 81.2 \\ \hline
SALT & \textbf{72.2} & \textbf{85.8/89.7} & \textbf{55.6} & \textbf{94.2} & \textbf{88.0/87.1} & \textbf{92.1} &\textbf{72.8/89.8} & \textbf{85.8/84.8} & \textbf{82.0} \\
\bottomrule
\end{tabular}
}
\vspace{-0.05in}
\caption{GLUE test set results on the GLUE evaluation server. All the methods fine-tune a pre-trained BERT\textsubscript{BASE} model. \textit{FreeAT} and \textit{SMART} did not report BERT\textsubscript{BASE} results in their paper or on the GLUE evaluation server. Model references: \textit{BERT\textsubscript{BASE}} \cite{devlin2018bert}, \textit{FreeLB} \cite{zhu2019freelb}.}
\label{tb:glue-test-results}
\end{table*}

\vspace{0.1in}
\noindent
\textbf{Implementation.}
Recall that to generate adversarial examples, we perturb the word embeddings. In NMT experiments, we perturb both the source-side and the target-side embeddings. This strategy is empirically demonstrated \cite{sano2019effective} to be more effective than perturbing only one side of the inputs.
We use \textit{Fairseq}\footnote{\url{https://github.com/pytorch/fairseq}} \cite{ott2019fairseq} to implement our algorithms.
We adopt the Transformer-base \cite{vaswani2017attention} architecture in all the low-resource experiments, except IWSLT'14 De-En. In this dataset, we use a model smaller than Transformer-base by decreasing the hidden dimension size from 2048 to 1024, and decreasing the number of heads from 8 to 4 (while dimension of each head doubles).
For the rich-resource experiments, we use the Transformer-big \cite{vaswani2017attention} architecture.
Training details are presented in Appendix~\ref{app:nmt}.

\vspace{0.1in}
\noindent
\textbf{Results.}
Experimental results for the low-resource experiments are summarized in Table~\ref{tb:low-resource}.
Notice that SMART, which utilizes conventional adversarial regularization, consistently outperforms standard adversarial training (Adv). Similar observations were also reported in \citet{miyato2016adversarial, sano2019effective}. This is because Adv generates perturbations using the correct examples, thus, the label information are ``leaked'' \cite{kurakin2016adversarial}.
Additionally, we can see that SALT is particularly effective in this low-resource setting, where it outperforms all the baselines by large margins. In comparison with the vanilla Transformer model, SALT achieves up to 2 BLEU score improvements on all the three datasets.

Table~\ref{tb:rich-resource} summarizes experiment results on the WMT'16 En-De dataset. We report the sacreBLEU~\cite{post2018call} score, which is a detokenzied version of the BLEU score that better reflects translation quality. We can see that SALT outperforms all the baseline methods by notable margins, and it improves upon the vanilla Transformer model by 1.2 BLEU score.

\begin{figure*}[t!]
    \centering
    \begin{subfigure}{0.33\textwidth}
        \centering
        \includegraphics[width=1.0\textwidth]{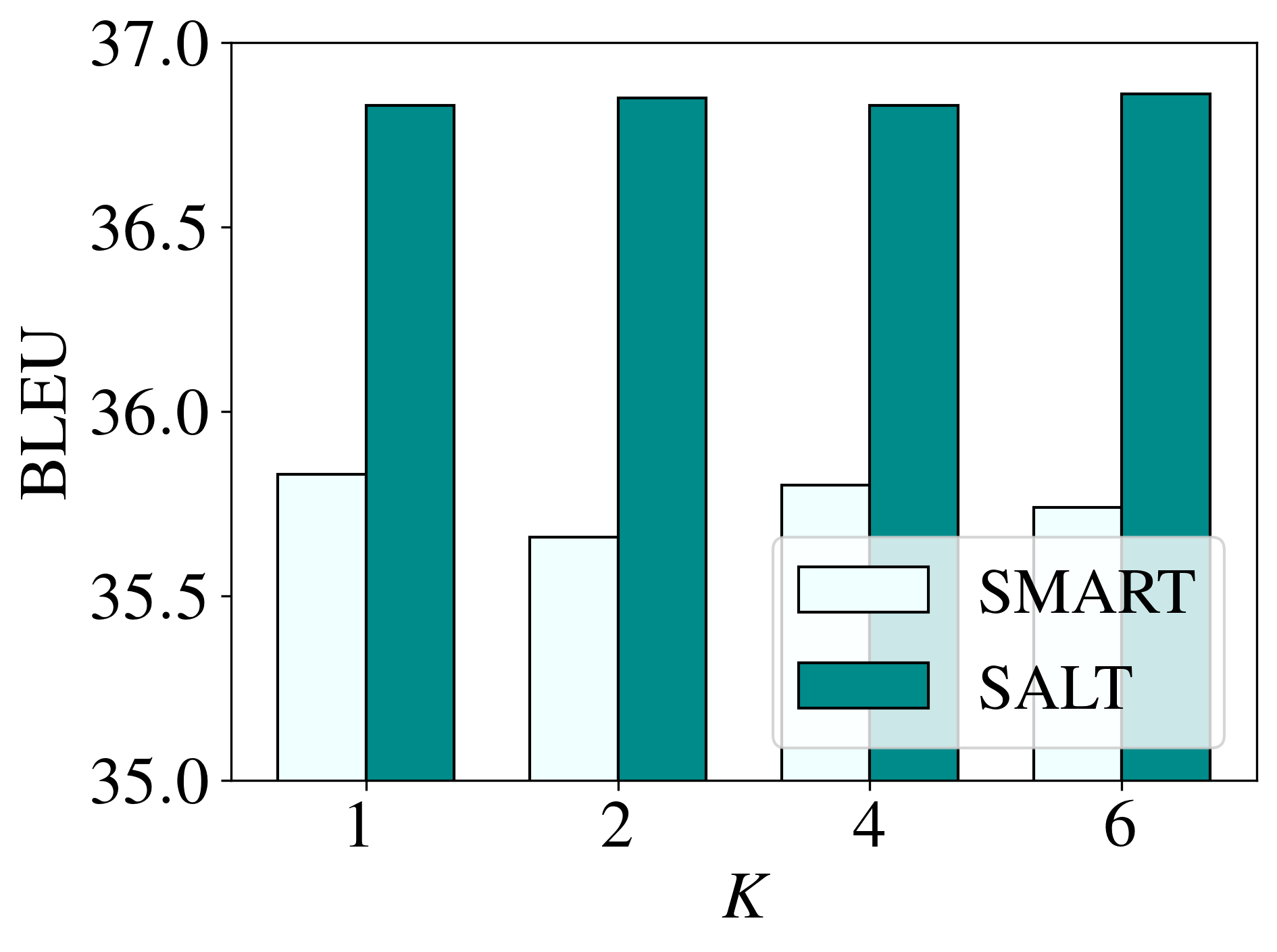}
        \vskip -0.1in
        \caption{Number of unrolling steps.}
        \label{fig:steps}
    \end{subfigure}%
    \begin{subfigure}{0.33\textwidth}
        \centering
        \includegraphics[width=1.0\linewidth]{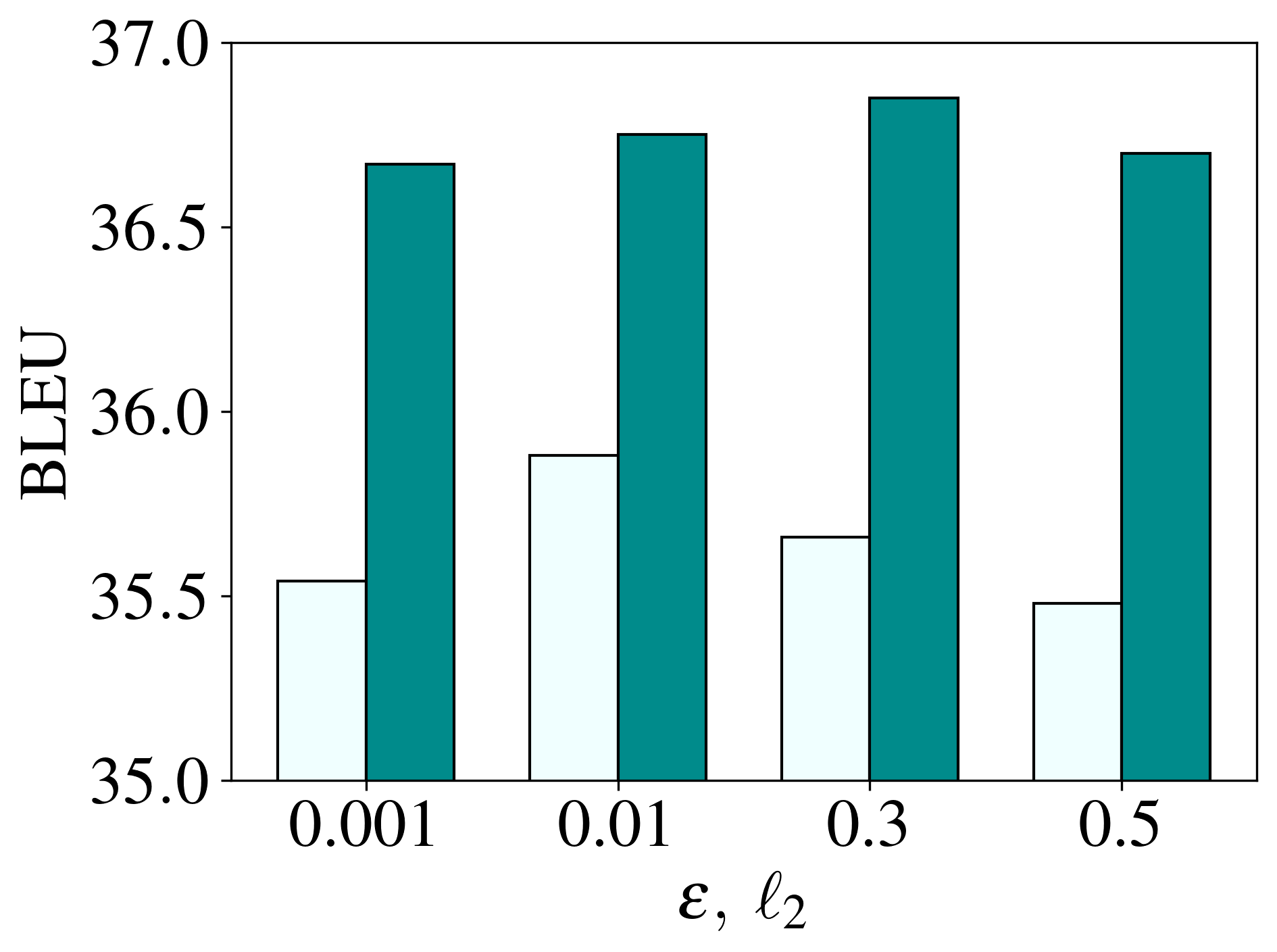}
        \vskip -0.1in
        \caption{Perturbation strength $\epsilon$, $\ell_2$ case.}
        \label{fig:eps}
    \end{subfigure}%
    \begin{subfigure}{0.33\textwidth}
        \centering
        \includegraphics[width=1.0\linewidth]{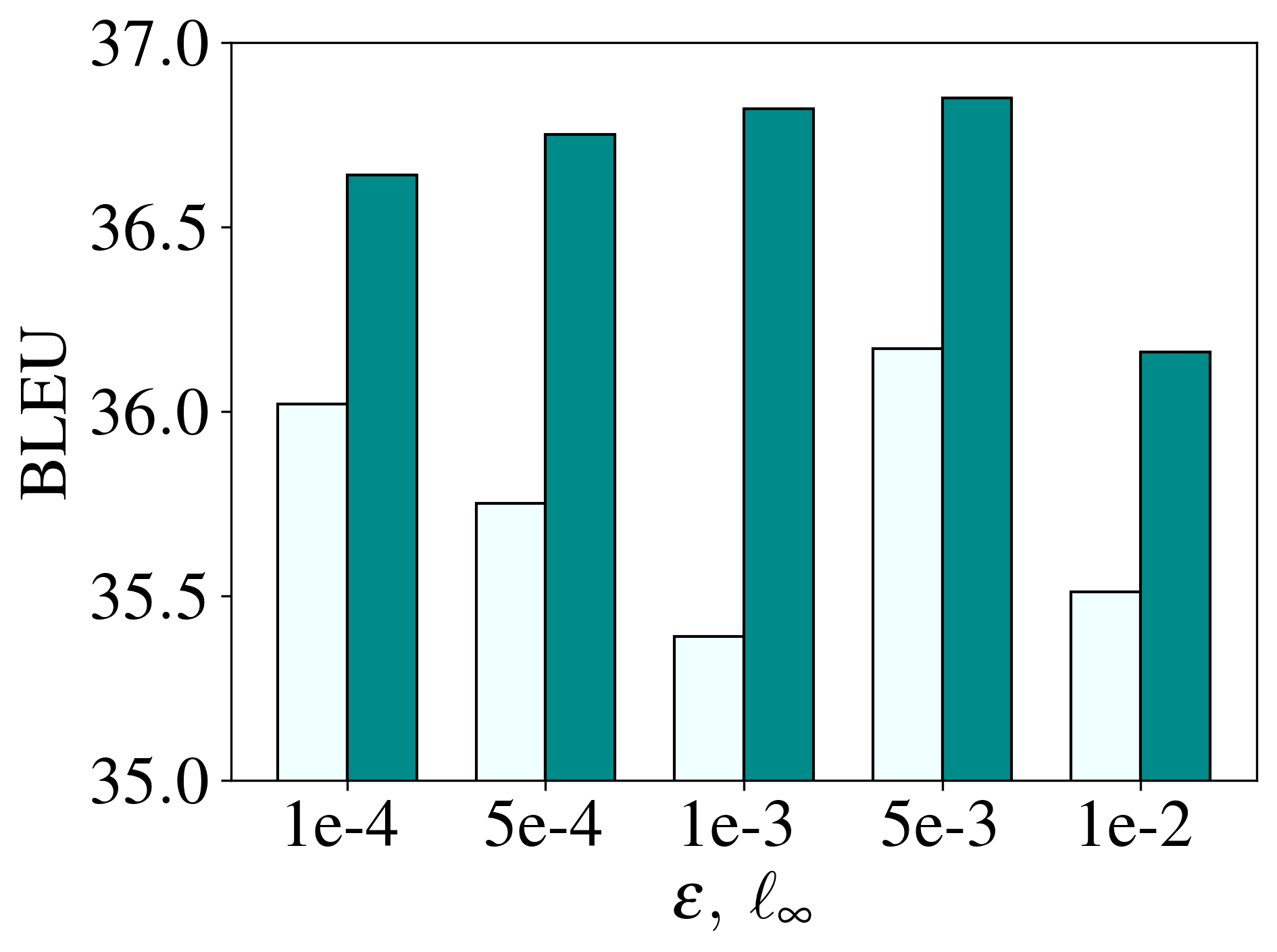}
        \vskip -0.1in
        \caption{Perturbation strength $\epsilon$, $\ell_\infty$ case.}
        \label{fig:linf}
    \end{subfigure}
    \vskip -0.05in
    \caption{Relation between BLEU score and different factors on the IWSLT'14 De-En dataset.}
\end{figure*}

\subsection{Natural Language Understanding}
\label{sec:exp-nlu}

\noindent \textbf{Datasets.}
We demonstrate the effectiveness of SALT on the General Language Understanding Evaluation (GLUE) benchmark \cite{wang2018glue}, which is a collection of nine NLU tasks. The benchmark includes question answering~\cite{squad1}, linguistic acceptability (CoLA, \citealt{cola2018}), sentiment analysis (SST, \citealt{sst2013}), text similarity (STS-B, \citealt{sts-b2017}), paraphrase detection (MRPC, \citealt{mrpc2005}), and natural language inference (RTE \& MNLI, \citealt{rte1,rte2,rte3,rte5,mnli2018}) tasks. Dataset details can be found in Table~\ref{tab:glue} (Appendix~\ref{app:nlu}).

\vspace{0.1in}
\noindent \textbf{Implementation.}
We evaluate our algorithm by fine-tuning a pre-trained BERT-base \cite{devlin2018bert} model. Our implementation is based on the \textit{MT-DNN} code-base \citep{liu2019multi,mtdnn2020demo}\footnote{\url{https://github.com/microsoft/MT-DNN}}.
Training details are presented in Appendix~\ref{app:nlu}.

\vspace{0.1in}
\noindent \textbf{Results.}
Table~\ref{tb:glue-results} summarizes experiment results on the GLUE development set. We can see that SALT outperforms BERT\textsubscript{BASE} in all the tasks. Further, our method is particularly effective for small datasets, such as RTE, MRPC, and CoLA, where we achieve 9.4, 4.3, and 6.3 absolute improvements, respectively.
Comparing with other adversarial training baselines, i.e., FreeAT, FreeLB, and SMART, our method achieves notable improvements in all the tasks.

We highlight that SALT achieves a 84.5 average score, which is significantly higher than that of the vanilla BERT\textsubscript{BASE} (+3.0) fine-tuning approach. Also, our average score is higher than the scores of baseline adversarial training methods (+1.9, +1.2, +0.7 for FreeAT, FreeLB, SMART, respectively).
Moreover, the 84.5 average score is even higher than fine-tuning BERT\textsubscript{LARGE} (+0.5), which contains three times more parameters than the backbone of SALT.

Table~\ref{tb:glue-test-results} summarizes results on the GLUE test set. We can see that SALT consistently outperforms BERT\textsubscript{BASE} and FreeLB across all the tasks.

\subsection{Parameter Study}

We highlight that SALT \textbf{does not introduce additional tuning parameter} comparing with conventional adversarial regularization approaches.

\vspace{0.05in}
\noindent $\diamond$ \textbf{Robustness to the number of unrolling steps.}
From Figure~\ref{fig:steps}, we can see that SALT is robust to the number of unrolling steps.
As such, setting the unrolling steps $K=1$ or $2$ suffices to build models that generalize well.

\vspace{0.05in}
\noindent $\diamond$ \textbf{Robustness to the perturbation strength.}
Unrolling is robust to the perturbation strength within a wide range, as indicated in Figure~\ref{fig:eps}.
Meanwhile, the performance of SMART consistently drops when we increase $\epsilon$ from 0.01 to 0.5.
This indicates that the unrolling algorithm can withstand stronger perturbations than conventional adversarial regularization approaches.

\vspace{0.05in}
\noindent $\diamond$ \textbf{$\ell_2$ constraints vs. $\ell_\infty$ constraints.} 
Figure~\ref{fig:linf} illustrates model performance with respect to different perturbation strength in the $\ell_\infty$ case. Notice that in comparison with the $\ell_2$ case (Figure~\ref{fig:eps}), SALT achieves the same level of performance, but the behavior of SMART is unstable.
Additionally, SALT is stable within a wider range of perturbation strength in the $\ell_2$ than in the $\ell_\infty$ case, which is the reason that we adopt $\ell_2$ constraints in the experiments.

\begin{figure*}[t!]
    \centering
    \begin{subfigure}{0.33\textwidth}
        \centering
        \includegraphics[width=1.0\textwidth]{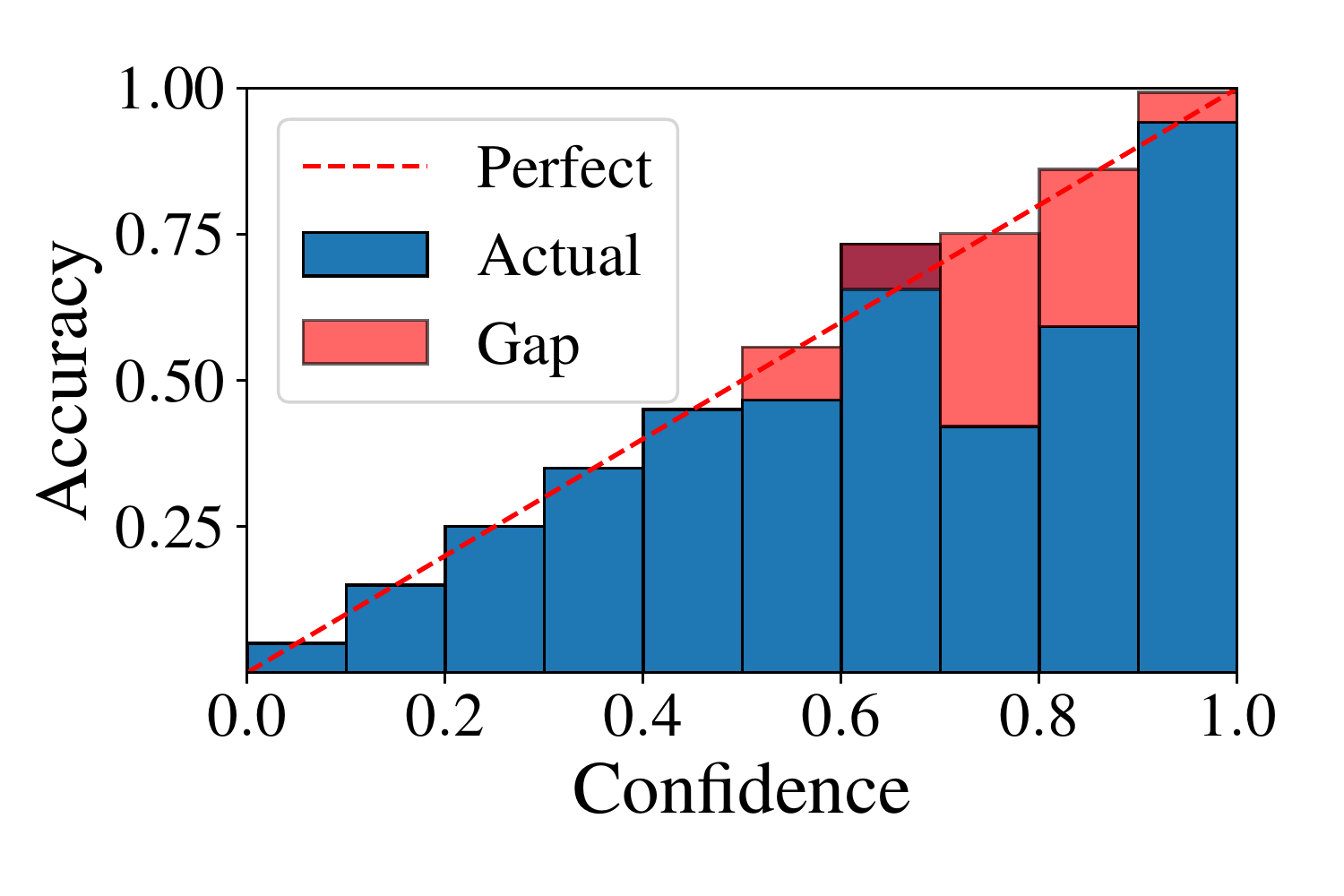}
        \vskip -0.1in
        \caption{BERT\textsubscript{BASE} (ECE: $6.09\%$).}
        \label{fig:cali_sst_base}
    \end{subfigure}%
    \begin{subfigure}{0.33\textwidth}
        \centering
        \includegraphics[width=1.0\linewidth]{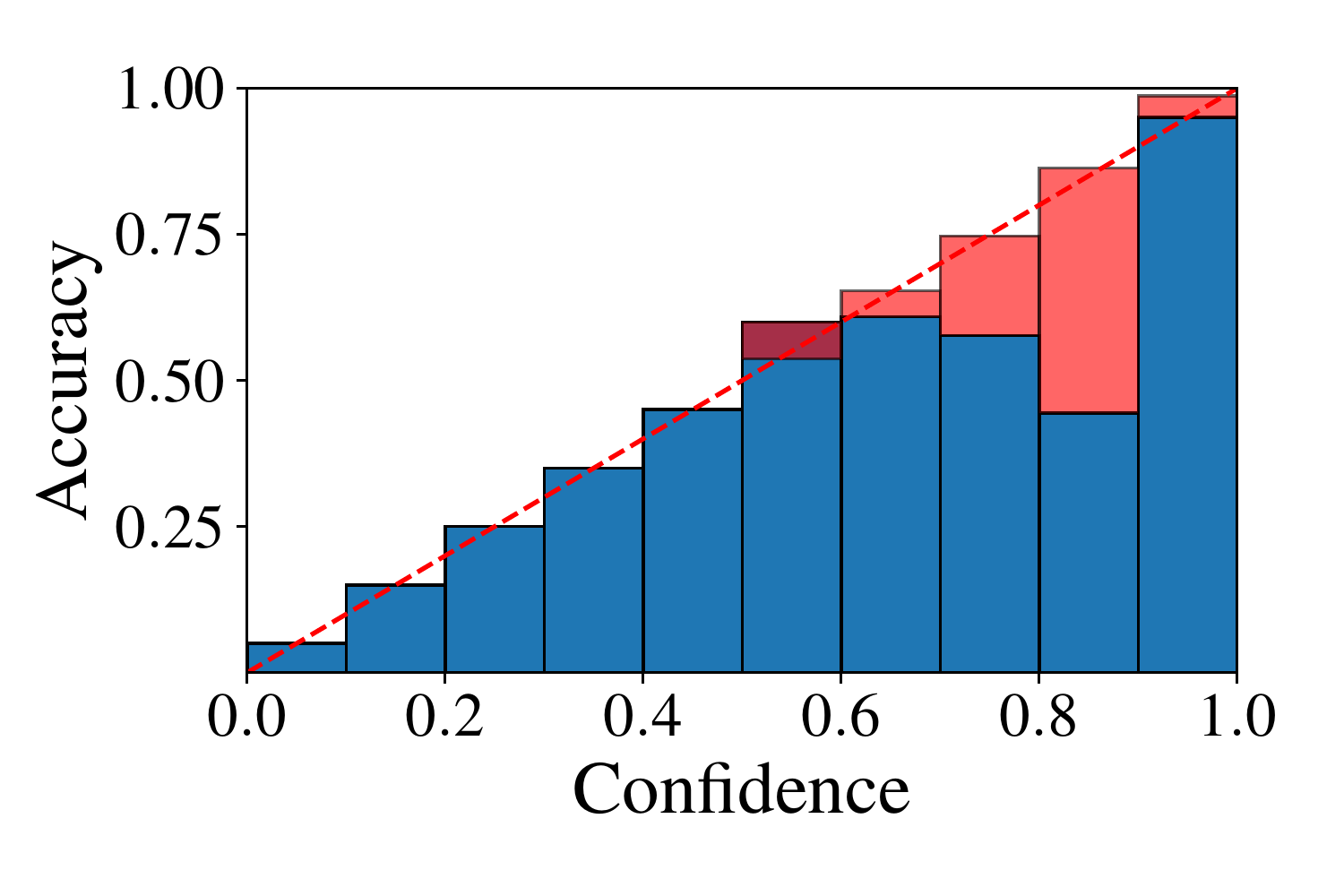}
        \vskip -0.1in
        \caption{SMART (ECE: $5.08\%$).}
        \label{fig:cali_sst_smart}
    \end{subfigure}%
    \begin{subfigure}{0.33\textwidth}
        \centering
        \includegraphics[width=1.0\linewidth]{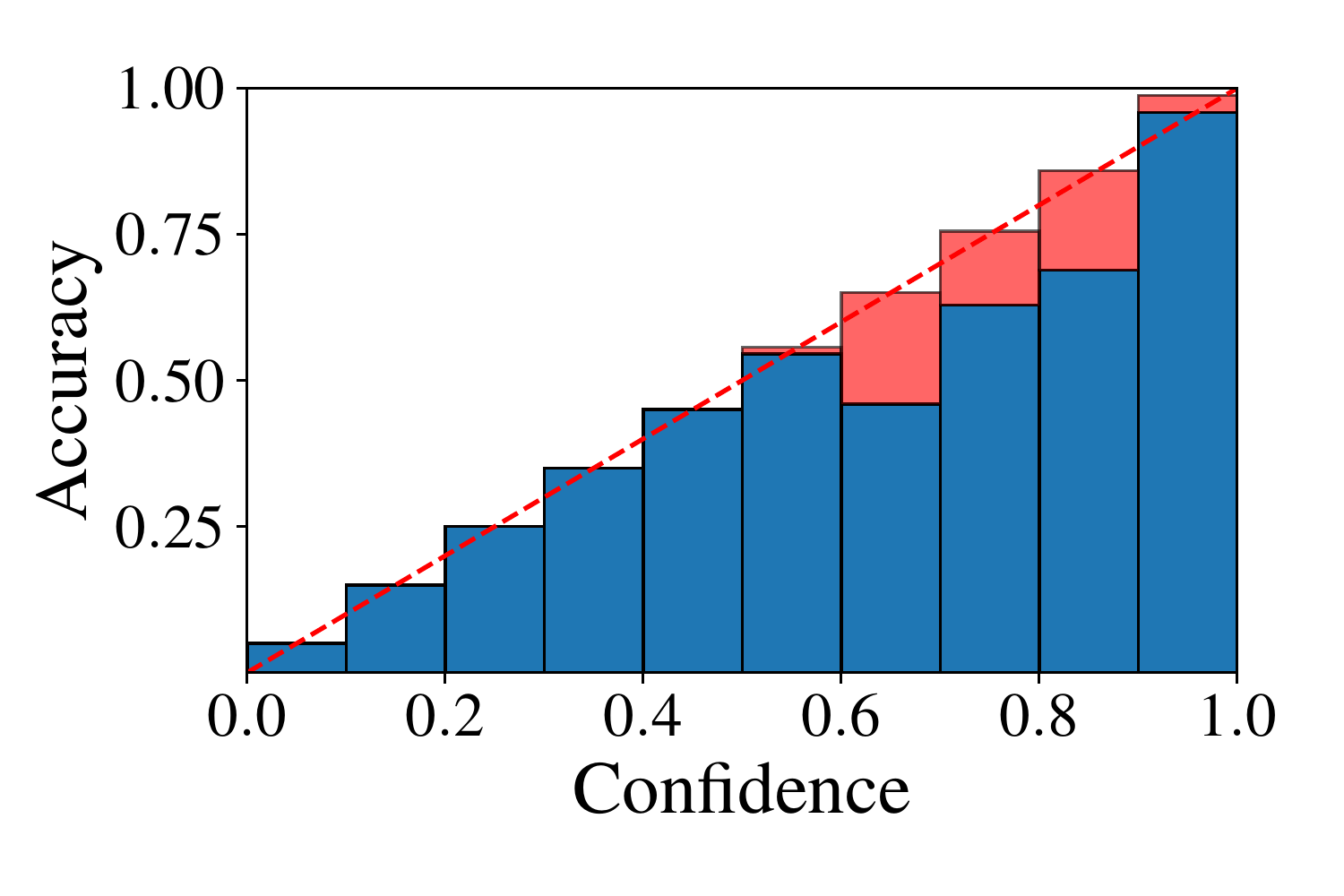}
        \vskip -0.1in
        \caption{SALT (ECE: $4.06\%$).}
        \label{fig:cali_sst_salt}
    \end{subfigure}
    \vskip -0.05in
    \caption{Reliability diagrams on SST. Perfect Calibration: confidence $=$ accuracy; ECE: the lower the better. }
    \label{fig:cali_sst}
\end{figure*}

\begin{figure}[!t]
    \centering
    \includegraphics[width=0.8\linewidth]{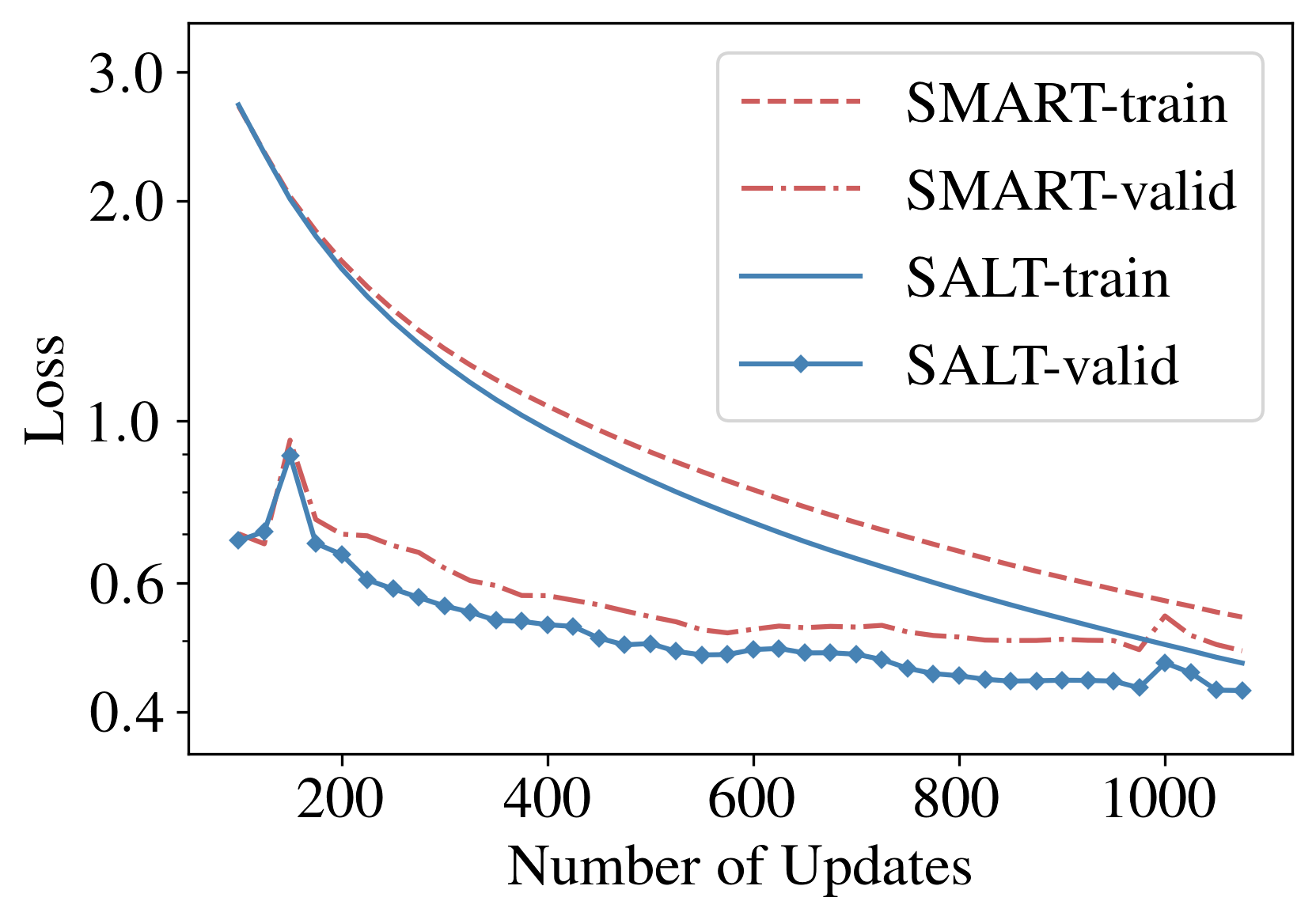}
    \includegraphics[width=0.8\linewidth]{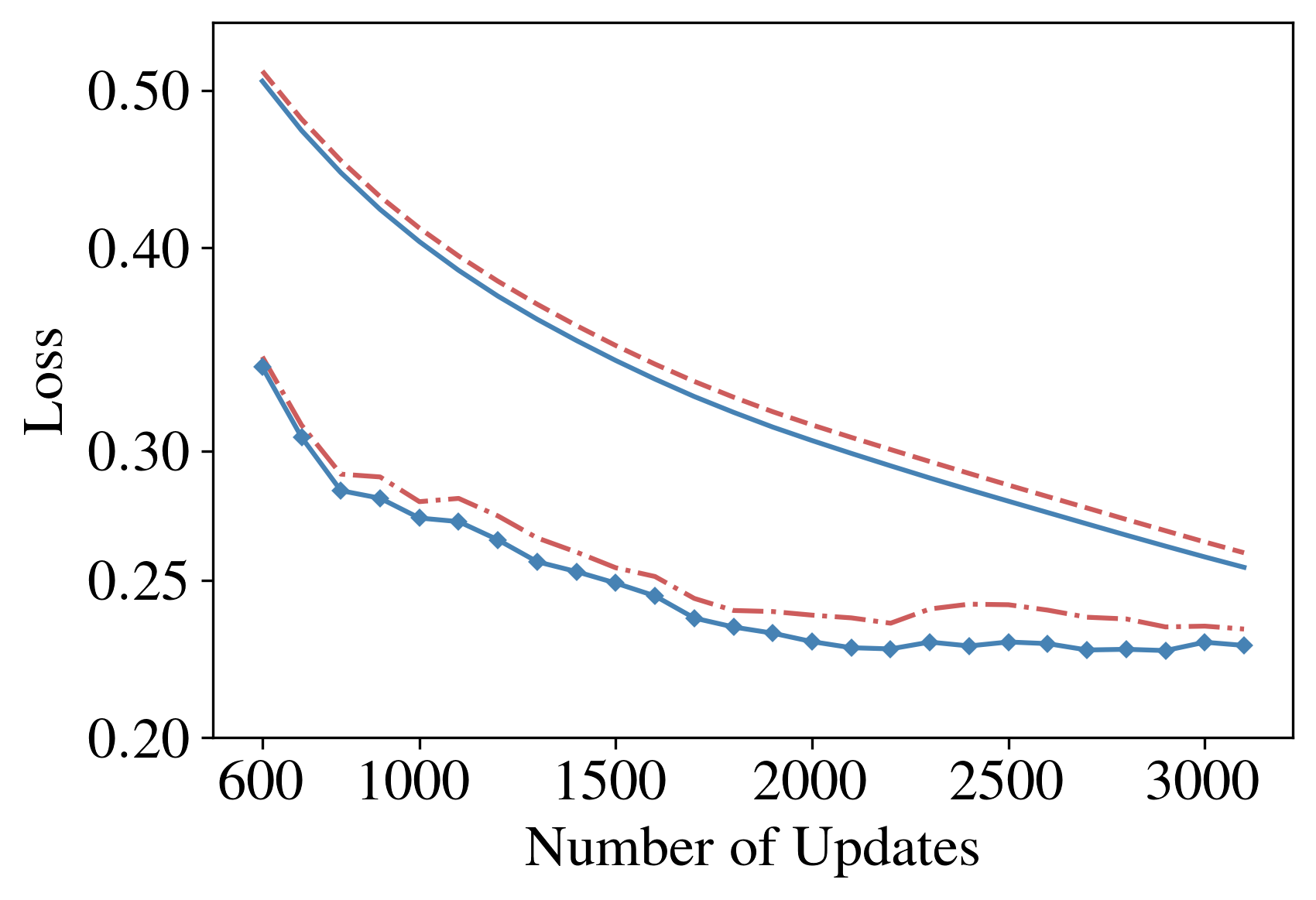}
    \vskip -0.1in
    \caption{Training and validation loss of \textit{SMART} and \textit{SALT} on STS-B (upper) and SST-2 (lower) datasets.}
    \label{fig:glue_loss}
\end{figure}

\subsection{Analysis}

\noindent $\diamond$ \textbf{Unrolling reduces bias.}
In Figure~\ref{fig:glue_loss}, we visualize the training and the validation error on the STS-B and the SST datasets from the GLUE benchmark.
As mentioned, conventional adversarial regularization suffers from over-strong perturbations, such that the model cannot fit the unperturbed data well.
This is supported by the fact that the training loss of SALT is smaller than that of SMART, which means SALT fits the data better.
SALT also yields a smaller loss than SMART on the validation data, indicating that the Stackelberg game-formulated model exhibits better generalization performance.

\vspace{0.05in}
\noindent $\diamond$ \textbf{Adversarial robustness.}
Even though the primary focus of SALT is model generalization, we still test its robustness on the Adversarial-NLI (ANLI, \citealt{nie2019adversarial}) dataset. The dataset contains 163k data, which are collected via a human-and-model-in-the-loop approach. From Table~\ref{tb:anli}, we can see that SALT improves model robustness upon conventional methods (i.e., SMART).

\begin{table}[!t]
\centering
\begin{tabular}{l|cccc}
\toprule
& \multicolumn{4}{c}{Dev} \\
& R1 & R2 & R3 & All \\ \midrule
BERT\textsubscript{BASE} & 53.3 & 43.0 & 44.7 & 46.8 \\
SMART & 54.1 & 44.4 & 45.3 & 47.8 \\ \hline
SALT & \textbf{56.6} & \textbf{46.2} & \textbf{45.9} & \textbf{49.3} \\ \bottomrule
& \multicolumn{4}{c}{Test} \\
& R1 & R2 & R3 & All \\ \midrule
BERT\textsubscript{BASE} & 54.1 & 44.9 & 46.6 & 48.4 \\
SMART & 54.3 & 46.4 & 46.5 & 48.9 \\ \hline
SALT & \textbf{55.4} & \textbf{47.7} & \textbf{46.7} & \textbf{49.7} \\ \bottomrule
\end{tabular}
\vspace{-0.05in}
\caption{Experimental results on the ANLI dataset. Model references: \textit{BERT\textsubscript{BASE}} \cite{devlin2018bert}, \textit{SMART} \cite{jiang2019smart}.}
\label{tb:anli}
\end{table}

\begin{figure}[t!]
    \centering
    \includegraphics[width=0.48\linewidth]{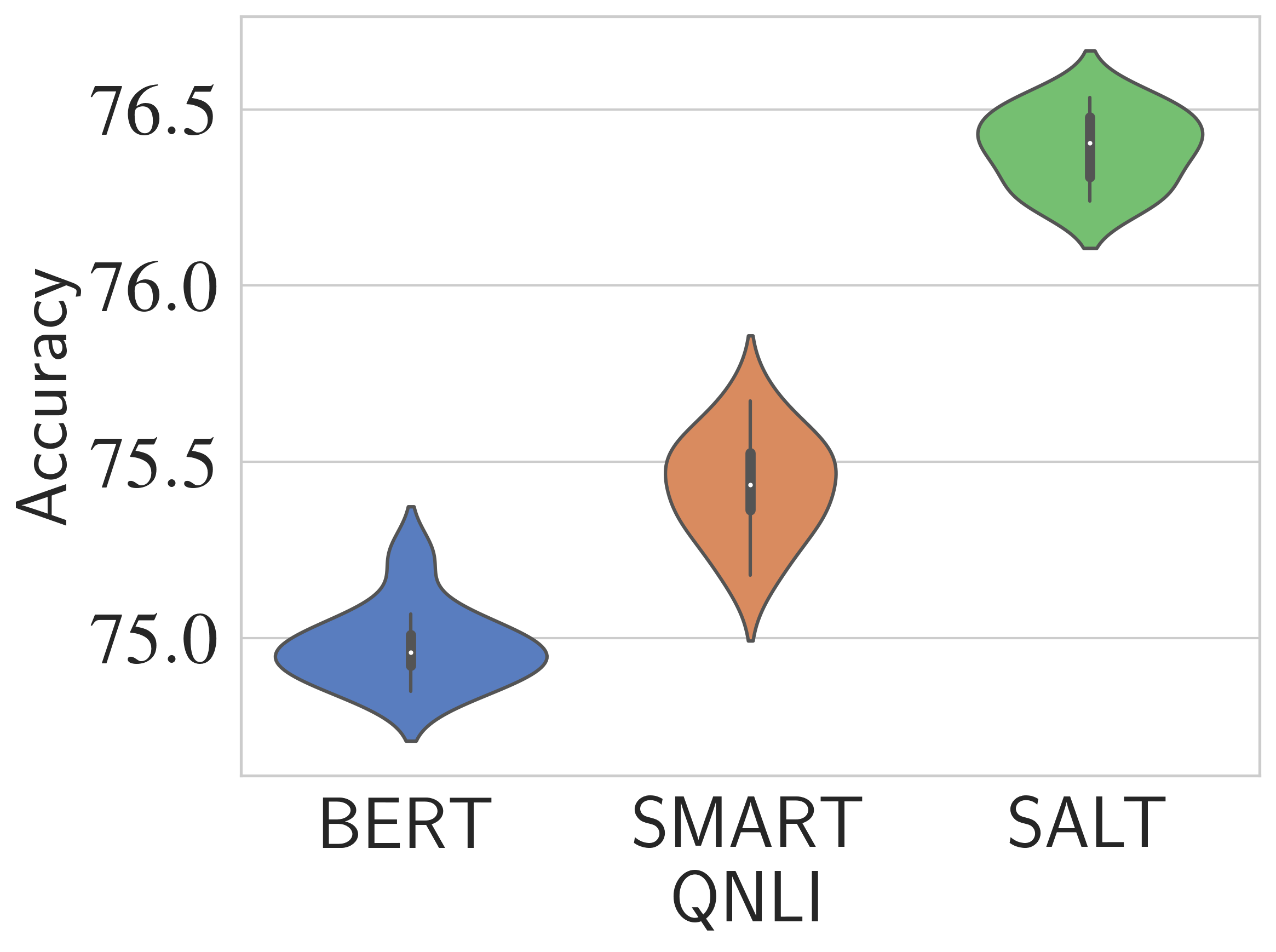}
    \includegraphics[width=0.48\linewidth]{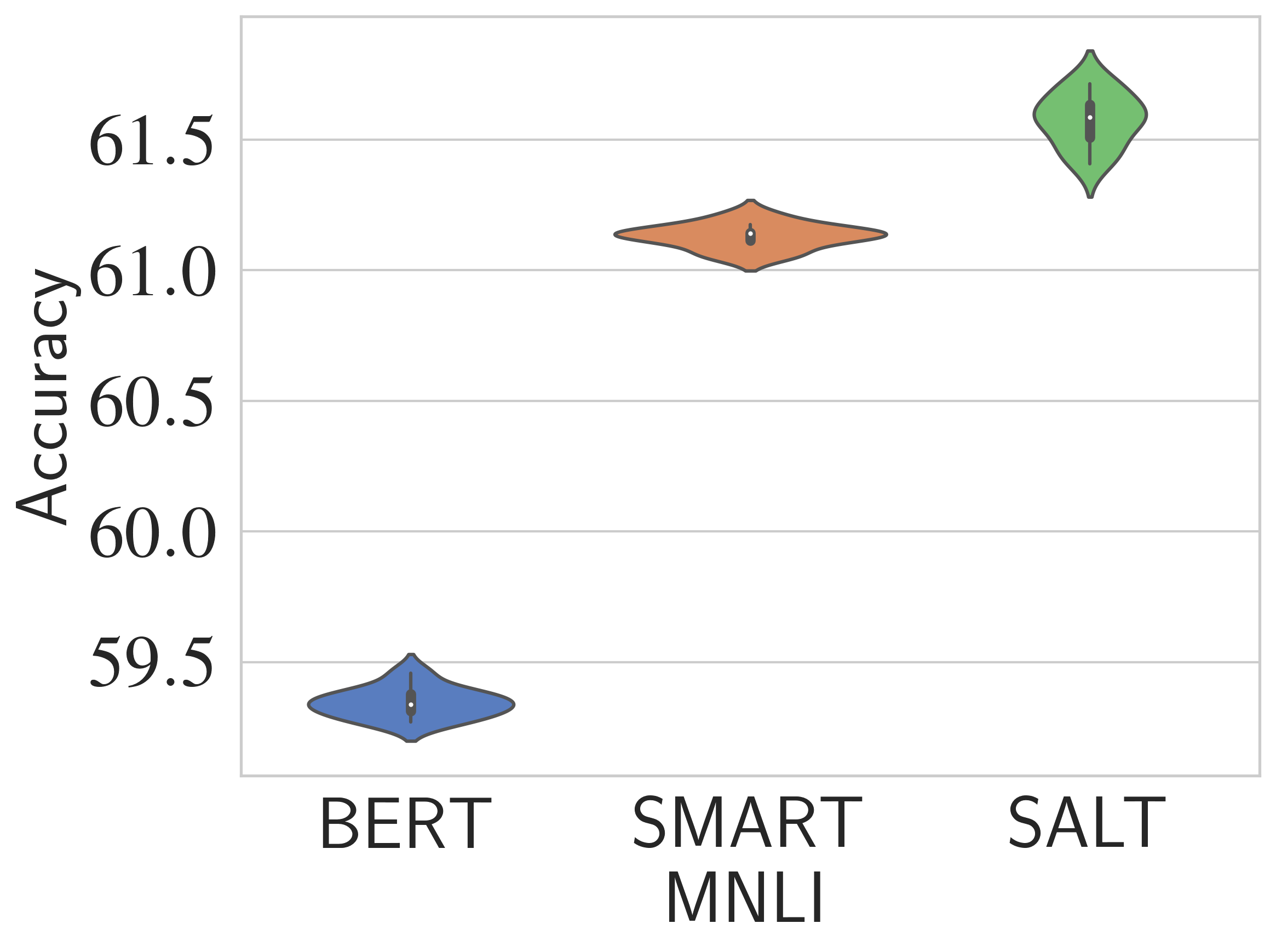}
    \vskip -0.1in
    \caption{Probing experiments. Each violin plot is based on 10 runs with different random seeds.}
    \label{fig:probing}
\end{figure}

\vspace{0.05in}
\noindent $\diamond$ \textbf{Probing experiments.}
For each method, we first fine-tune a BERT\textsubscript{BASE} model on the SST-2 dataset. Then, we only tune a prediction head on other datasets while keeping the representations fixed. Such a method directly measures the quality of representations generated by different models. As illustrated in Fig.~\ref{fig:probing}, SALT outperforms the baseline methods by large margins.

\vspace{0.05in}
\noindent $\diamond$ \textbf{Classification Model Calibration.} 
Adversarial regularization also helps model calibration \citep{stutz2020confidence}. A well-calibrated model produces reliable confidence estimation (i.e., confidence $\simeq$ actual accuracy), where the confidence is defined as the maximum output probability calculated by the model. We evaluate the calibration performance of BERT\textsubscript{BASE}, SMART, and SALT by the Expected Calibration Error (ECE, \citealt{niculescu2005predicting}). We plot the reliability diagram (confidence vs. accuracy) on the SST task in Fig.~\ref{fig:cali_sst} (see Appendix~\ref{app:calibration} for details). As we can see, BERT\textsubscript{BASE} and SMART are more likely to make overconfident predictions. SALT reduces ECE, and its corresponding reliability diagram aligns better with the perfect calibration curve.

\vspace{0.05in}
\noindent $\diamond$ \textbf{Comparison with Unrolled-GAN.}
The unrolling technique has been applied to train GANs (Unrolled-GAN, \citealt{metz2016unrolled}). However, subsequent works find that this approach not necessarily improves training \cite{grnarova2017online, tran2019self, doan2019line}.
This is because Unrolled-GAN unrolls its discriminator, which has a significant amount of parameters. Consequently, the unrolling algorithm operates on a very large space, rendering the stochastic gradients that are used for updating the discriminator considerably noisy.
In SALT, the unrolling space is the sample embedding space, the dimension of which is much smaller than the unrolling space of GANs. Therefore, unrolling is more effective for NLP tasks.

\section{Conclusion}

We propose SALT, an adversarial regularization method that employs a Stackelberg game formulation. Such a formulation induces a competition between a leader (the model) and a follower (the adversary).
In SALT, the leader is in an advantageous position by recognizing the follower's strategy, and this strategic information is captured by the Stackelberg gradient. We compute the Stackelberg gradient, and hence find the equilibrium of the Stackelberg game, using an unrolled optimization approach.
Empirical results NMT and NLU tasks suggest the superiority of SALT to existing adversarial regularization methods.

\bibliography{anthology,custom}
\bibliographystyle{acl_natbib}

\clearpage
\appendix 
\section{Virtual Adversarial Training}
\label{app:vat}

Virtual adversarial training (VAT, \citealt{miyato2016adversarial}) solves the following min-max optimization problem:
\begin{align*}
    \min_{\theta} \cF(\theta, \delta^*) &= \cL(\theta) + 
    \frac{\alpha}{n} \sum_{i=1}^n \ell_{v}( x_i, \delta_i^*, \theta ) , \\
    \delta^*_i &= \argmax_{\norm{\delta_i} \leq \epsilon} \ell_{v}( x_i, \delta_i, \theta ),
\end{align*}
where
\begin{equation*}
    \ell_{v}(x_i, \delta_i, \theta) = \mathrm{KL}\big( f(x_i, \theta) ~||~ f(x_i + \delta_i, \theta) \big).
\end{equation*}
Note that the objective of the minimization problem is a function of both the model parameters and the perturbations.

Because the min problem and the max problem are operating on the same loss function, i.e., the min problem seeks to minimize $\ell_v$, while the max problem tries to maximize $\ell_v$, this min-max optimization is essentially a zero-sum game. And we can find the game's equilibrium using gradient descent/ascent algorithms.

Specifically, the adversarial player first generate an initial perturbation $\delta^0$, and then refines it using $K$ steps of projected gradient ascent, i.e.,
\begin{align*}
    &\delta^k = \Pi_{\norm{\cdot} \leq \epsilon} \left(\delta^{k-1} + \eta \frac{\partial \ell_v(x, \delta^{k-1}, \theta)}{\partial \delta^{k-1}} \right), \\
    &\quad \text{for } k = 1, \cdots, K.
\end{align*}
Here $\Pi$ denotes projection onto the $\ell_2$-ball or the $\ell_\infty$-ball. Empirically, we find that these two choices yield very similar performance, although adversarial training models is robust to $\epsilon$ within a wider range when applying the $\ell_2$ constraint.

After obtaining the $K$-step refined perturbation $\delta^K$, we use gradient descent to update the model parameters $\theta$. Concretely, the gradient of the model parameters is computed as
\begin{equation} \label{eq:vat-gradient}
    \frac{\partial \cF(\theta, \delta^K)}{\partial \theta}
    = \frac{\dd \ell(f(x_i,\theta), y_i)}{\dd \theta} 
    + \alpha \frac{\partial \ell_v(x, \delta^K, \theta)}{\partial \theta}.
\end{equation}
The training algorithm is demonstrated in Algorithm~\ref{alg:vat}.

Note that in this paper, we target for models' generalization performance on the unperturbed test data, therefore we do not want a strong adversary that ``traps'' the model parameters to a bad local optima. Most of the existing algorithms achieve this goal by carefully tuning the hyper-parameters $\epsilon$ and $K$, i.e., a small $\epsilon$ usually generates weaker adversaries, so does a small $K$.
However, these heuristics do not work well, and at times $\delta^K$ is too strong. Consequently, conventional adversarial training results in undesirable underfitting on the clean data.

\begin{algorithm}[!htb]
\caption{Virtual Adversarial Training.}
\label{alg:vat}
\KwIn{$\cD$: dataset; $T$: total number of training iterations; $\sigma^2$: variance of initial perturbations; $K$: number of inner training iterations; $\eta$: step size to update $\delta$; Optimizer: optimizer to update $\theta$.}
\textbf{Initialize:} model parameters $\theta$\;
\For{$t = 1, \cdots, T$}{
\For{$(x,y) \in \cD$}{
    Initialize $\delta^0 \sim \cN(0, \sigma^2 I)$\;
    \For{$k = 1, \cdots, K$}{
        $g^k \leftarrow \partial \ell_v(x_i, \delta_i, \theta) / \partial \delta_i$\;
        $\delta^k \leftarrow \Pi( \delta^{k-1} + \eta g^k )$\;
    }
    Compute the gradient $g_\theta$ using Eq.~\ref{eq:vat-gradient}\;
    $\theta \leftarrow \text{Optimizer}(g_\theta)$\;
}
}
\KwOut{$\theta$}
\end{algorithm}

\begin{table*}[!t]
	\begin{center}
		\begin{tabular}{l|l|c|c|c|c|c}
			\toprule 
			\bf Corpus &Task& \#Train & \#Dev & \#Test   & \#Label &Metrics\\ \midrule
			\multicolumn{6}{@{\hskip1pt}r@{\hskip1pt}}{Single-Sentence Classification (GLUE)} \\ \hline
			CoLA & Acceptability&8.5k & 1k & 1k & 2 & Matthews corr\\ \hline
			SST & Sentiment&67k & 872 & 1.8k & 2 & Accuracy\\ \midrule
			\multicolumn{6}{@{\hskip1pt}r@{\hskip1pt}}{Pairwise Text Classification (GLUE)} \\ \hline
			MNLI & NLI& 393k& 20k & 20k& 3 & Accuracy\\ \hline
            RTE & NLI &2.5k & 276 & 3k & 2 & Accuracy \\ \hline
			QQP & Paraphrase&364k & 40k & 391k& 2 & Accuracy/F1\\ \hline
            MRPC & Paraphrase &3.7k & 408 & 1.7k& 2&Accuracy/F1\\ \hline
			QNLI & QA/NLI& 108k &5.7k&5.7k&2& Accuracy\\ \midrule
			\multicolumn{5}{@{\hskip1pt}r@{\hskip1pt}}{Text Similarity (GLUE)} \\ \hline
			STS-B & Similarity &7k &1.5k& 1.4k &1 & Pearson/Spearman corr\\ \bottomrule
		\end{tabular}
	\end{center}
	\vskip -0.05in
	\caption{Summary of the GLUE benchmark.}
	\label{tab:glue}
\end{table*}

\section{Training Details}

\subsection{Neural Machine Translation}
\label{app:nmt}

\begin{table*}[t!]
\centering
\begin{tabular}{l|cccccc|cc}
\toprule
&  Batch & lr\textsubscript{leader} & lr\textsubscript{follower} & $\sigma$ & $\epsilon$ & $K$ & Beam & Len-Pen \\ \midrule
En-Vi (IWSLT'15) & $64k$ & $1 \times 10^{-3}$ & $1 \times 10^{-5}$ & $1 \times 10^{-4}$ & $0.1$ & $1$ & $10$ & $1.0$ \\
De-En (IWSLT'14) & $64k$ & $1 \times 10^{-3}$ & $1 \times 10^{-4}$ & $1 \times 10^{-4}$ & $0.3$ & $1$ & $9$ & $1.5$ \\
Fr-En (IWSLT'16) & $64k$ & $1 \times 10^{-3}$ & $1 \times 10^{-5}$ & $1 \times 10^{-5}$ & $0.3$ & $1$ & $10$ & $2.0$ \\ \hline
En-De (WMT'16) & $450k$ & $1 \times 10^{-3}$ & $1 \times 10^{-4}$ & $1 \times 10^{-4}$ & $0.3$ & $1$ & $4$ & $0.6$ \\
\bottomrule
\end{tabular}
\vskip -0.05in
\caption{Hyper-parameters for machine translation. Here, $\sigma$ is the standard deviation of the initial perturbations, $\epsilon$ is the perturbation strength, $K$ is the number of unrolling steps, \textit{Beam} is the size of beam search, and \textit{Len-Pen} is the length penalty parameter during beam search.}
\label{tab:mt-parameter}
\end{table*}

For the rich-resource WMT'16 En-De dataset, we use the pre-processed data from \citet{ott2018scaling}\footnote{\url{https://github.com/pytorch/fairseq/tree/master/examples/scaling_nmt}}. 
For the low-resource datasets, we use byte-pair encoding \cite{sennrich2015neural} with 10,000 merge operations to build the vocabulary for the IWSLT ('14, '15, '16) datasets. We follow the scripts in \citet{ott2019fairseq}\footnote{\url{https://github.com/pytorch/fairseq/tree/master/examples/translation}\label{footnote:nmt}} for other pre-processing steps.

We use Adam \cite{kingma2014adam} as the leader's (i.e., the upper level problem that solves for model parameters) optimizer, and we set $\beta=(0.9,0.98)$. 
The follower's (i.e., the lower level problem that solves for perturbations) optimizer is chosen from Adam and SGD, where we observe only marginal empirical differences between these two choices.
For low-resource translation, we set the batch size to be equivalent to 64k tokens.
For example, when running the experiments on 4 GPUs, we set the tokens-per-GPU to be 8,000, and we accumulate gradients for 2 steps. For rich-resource translation, we set the batch size to be equivalent to 450k tokens.
In all the experiments, we constrain each perturbation according to its sentence-level $\ell_2$ norm, i.e., $\norm{\delta}_2 \leq \epsilon$.
Other hyper-parameters are specified in Table~\ref{tab:mt-parameter}.


\subsection{Natural Language Understanding}
\label{app:nlu}

Details of the GLUE benchmark, including tasks, statistics, and evaluation metrics, are summarized in Table~\ref{tab:glue}.

We use Adam as both the leader's and the follower's optimizer, and we set $\beta = (0.9, 0.98)$. The learning rate of the leader $\text{lr}_{\text{leader}}$ is chosen from $\{5\times 10^{-5}, 1\times 10^{-4}, 5\times10^{-4}\}$, and the follower's learning rate is chosen from $\{1\times 10^{-5}, \text{lr}_{\text{leader}}\}$. We choose the batch size from $\{4, 8, 16, 32\}$, and we train for a maximum $6$ epochs with early-stopping based on the results on the development set. We apply a gradient norm clipping of $1.0$. We set the dropout rate in task specific layers to $0.1$. We choose standard deviation of initial perturbations $\sigma$ from $\{1\times10^{-5}, 1\times10^{-4}\}$, and $\ell_2$ constraints with perturbation strength $\epsilon=1.0$ are applied. We set the unrolling steps $K=2$. We report the best performance on each dataset individually.

\section{Model Calibration}
\label{app:calibration}

Many applications require trustworthy predictions that need to be not only accurate
but also well calibrated \citep{kong2020calibrated}.
A well-calibrated model is expected to output prediction
confidence comparable to its classification accuracy. For example, given 100
data points with their prediction confidence $0.6$, we expect $60$ of them to be
correctly classified. More precisely, for a data point $X$, we denote by $Y(X)$
the ground truth label, ${\hat Y}(X)$ the label predicted by the model, and $\hat{P}(X)$
  the output probability associated with the predicted label. The calibration error of the predictive model for a given confidence $p\in(0,1)$ is defined as:
\begin{equation}
    \mathcal{E}_p= \left|\mathbb{P}\left[\hat{Y}(X)=Y(X)|\hat{P}(X)=p\right] - p \right|.
    \label{eq:id}
\end{equation}
Since Eq.~\ref{eq:id} involves population quantities,
we usually adopt empirical approximations \citep{guo2017calibration} to estimate the calibration error. Specifically, we partition all data points into 10 bins of equal size according to their prediction confidence. Let $\mathcal{B}_m$ denote the bin with prediction confidence bounded between $\ell_m$ and $u_m$. Then, for any $p\in[\ell_m,u_m)$, we define the empirical calibration error as:
\begin{equation}
\hat{\mathcal{E}}_p=\hat{\mathcal{E}}_m=\frac{1}{|\mathcal{B}_m|}\Big|\sum_{i\in\mathcal{B}_m}\big[\mathbf{1}(\hat{y}_i=y_i)-\hat{p}_i\big]\Big|,
\end{equation}
where $y_i$, $\hat{y}_i$ and $\hat{p}_i$ are the true label, predicted label and confidence for sample $i$.

\textbf{Reliability Diagram} is a bar plot that compares $\hat{\mathcal{E}}_p$ against each bin, i.e., $p$. A perfectly calibrated would have $\hat{\mathcal{E}}_p = (\ell_m+u_m)/2$ for each bin.

\textbf{Expected Calibration Error (ECE)} is the weighted average of the calibration errors of all bins \citep{naeini2015obtaining} defined as:
\begin{align}
    {\rm ECE} =\sum_{m=1}^M\frac{|\mathcal{B}_m|}{n} \hat{\mathcal{E}}_{m},
    \label{eq:ece}
\end{align}
where $n$ is the sample size.

We remark that the goal of calibration is to minimize the calibration error without significantly sacrificing prediction accuracy. Otherwise, a random guess classifier can achieve zero calibration error.

\end{document}